%% file: main.tex
\documentclass[10pt,twocolumn,letterpaper]{article}

\usepackage[pagenumbers]{cvpr} 
\input{preamble}
\definecolor{cvprblue}{rgb}{0.21,0.49,0.74}
\usepackage[pagebackref,breaklinks,colorlinks,allcolors=cvprblue]{hyperref}
\usepackage{booktabs}
\usepackage{tabularx}
\usepackage{amssymb}
\usepackage{array}
\usepackage{amsmath}
\usepackage{makecell}
\usepackage{graphicx}
\usepackage{multirow}
\usepackage[table]{xcolor}
\usepackage{siunitx}

\definecolor{cartgray}{gray}{0.975}
\definecolor{lossgray}{gray}{0.945}

\def\confName{CVPR}
\def\confYear{2027}

\title{SPEM: Scan-Prior Entropy Modeling for LiDAR Point Cloud Compression}

\author{
Bojun Liu \quad
Yangzhi Ma \quad
Jie Li \quad
Dong Liu \quad
Li Li\\[2pt]
University of Science and Technology of China\\[3pt]
{\tt\small
\{liubj,mayz,lijl\}@mail.ustc.edu.cn}\\[1pt]
{\tt\small
\{dongeliu,lil1\}@ustc.edu.cn}
}

\begin{document}
\maketitle
\input{sec/0_abstract}    
\input{sec/1_intro}
\input{sec/2_Related_work}

\input{sec/3_Method}
\input{sec/4_Experiment}

\input{sec/5_Conclusion}

\section*{LLM Usage Statement}
In preparing this manuscript, we used ChatGPT and Codex to assist with academic writing and code development, including language refinement, manuscript organization, code implementation, and debugging. All core scientific ideas, methodological designs, experimental decisions, and interpretations of results were developed by the authors. All AI-assisted content and code were reviewed and validated by the authors, who take full responsibility for this work.

{
    \small
    \bibliographystyle{ieeenat_fullname}
    \bibliography{main}
}


\end{document}

%% file: sec/0_abstract.tex
\begin{abstract}
LiDAR point clouds exhibit scan-induced spatial regularities that are informative for geometry compression. Existing scan-aware approaches often exploit these regularities by transforming coded geometry into spherical or beam-aligned coordinate systems. However, this couples acquisition priors with geometry representation, altering spatial partitioning and quantization, and potentially preventing exact recovery of the original Cartesian geometry. We argue that acquisition priors should guide probability modeling rather than dictate the coding representation. We propose SPEM, a Scan-Prior Entropy Modeling framework that exploits LiDAR scanning regularities within the entropy model without requiring coordinate transformation. We introduce a multi-frequency spherical rotary positional encoding that captures scan-aligned spatial relationships in both backbone self-attention and decoded-context retrieval. Building upon this encoding, we develop a decoded-memory autoregressive predictor that efficiently exploits previously decoded occupancy information through reusable key/value caches. Experiments demonstrate state-of-the-art performance in both lossy and lossless LiDAR geometry compression, highlighting the effectiveness of scan-prior modeling across different coordinate representations.

\end{abstract}

%% file: sec/1_intro.tex
\section{Introduction}
\label{sec:intro}

LiDAR has become a fundamental sensing modality for autonomous driving, robotics, and large-scale 3D scene understanding~\cite{behley2019iccv,caesar2020nuscenes,roriz2021automotive,sun2020scalability},
providing accurate geometric measurements of the surrounding environment. With the rapid development of high-resolution and high-frequency LiDAR sensors, modern systems can capture increasingly dense and precise 3D observations, supporting a wide range of tasks such as localization, perception, mapping, and scene reconstruction.
However, continuously acquired LiDAR point clouds contain massive numbers of 3D points with high coordinate precision, imposing substantial storage and transmission overhead~\cite{schwarz2018emerging,graziosi2020overview, gao2025deep}.
Efficient point cloud geometry compression is therefore essential for reducing this cost while preserving geometric fidelity.

Unlike regularly images and videos, point clouds are inherently unordered and irregular. Existing compression methods therefore introduce structured representations
to organize point geometry and exploit its spatial or acquisition characteristics, including octrees~\cite{huang2020octsqueeze,fu2022octattention},
predtree~\cite{sun2026lpcm, sun2026inter}, one-chain structures~\cite{lasserre2021lidar}, range images~\cite{wang2026d,you2026bitric}, and sparse tensors~\cite{wang2021multiscale,wang2022sparse}.
Beyond generic geometric structures, recent methods further exploit the
acquisition prior of rotating LiDAR.
For example, SCP~\cite{luo2024scp} transforms Cartesian coordinates into spherical coordinates to expose the scanning pattern, while
PACE~\cite{zhu2026pace} further incorporates LiDAR intrinsic information through a cylindrical-beam representation. Both representations are organized into octree structures for entropy modeling, leading to improved compression performance. However, these approaches rely on scan-aligned coordinate transformations, which alter spatial partitioning and quantization and may prevent exact recovery of the original Cartesian geometry. More fundamentally, this raises a question: \emph{can LiDAR scanning priors be exploited directly within the entropy model, without requiring a specific geometry representation?}

Our key insight is that LiDAR scanning priors are reflected in the spatial relationships of points, rather than tied to a particular coordinate representation. Even in a Cartesian octree, node locations retain the range and angular relationships induced by the LiDAR acquisition process. Based on this observation, we propose SPEM, a Scan-Prior Entropy Modeling framework that incorporates these scan-aware relationships into occupancy probability modeling without requiring coordinate transformation. This decoupling preserves the original geometry for strict lossless coding while enabling effective entropy modeling for both lossy and lossless compression.

Specifically, SPEM contains two key designs.
First, we introduce a spherical positional encoding that maps the physical positions of octree tokens into spherical coordinates $(r,\theta,\phi)$ for attention computation, without altering the underlying Cartesian octree.
Multi-frequency rotary encoding is then applied to the attention queries and keys, allowing the model to capture scan-aligned spatial dependencies while leaving the coded Cartesian geometry unchanged. Second, we develop a stage-scalable decoded-memory predictor, where the current tokens query previously decoded occupancy information through cross-attention. The decoded tokens are encoded into reusable key/value caches and progressively
accumulated across prediction stages, enabling effective autoregressive context modeling without repeatedly recomputing historical representations. Together, these designs allow SPEM to explicitly preserve Cartesian geometry while implicitly exploiting the LiDAR scanning prior for both lossy and lossless point cloud compression.

In summary, the proposed SPEM offers the following key contributions:

\begin{itemize}
    \item We introduce a spherical rotary positional encoding that injects the
    LiDAR scanning prior directly into the entropy model, enabling scan-aware
    spatial modeling without requiring explicit coordinate transformation.

    \item Building upon the proposed positional encoding, we develop a
    cross-attention-based autoregressive probability predictor that efficiently
    exploits previously decoded occupancy information through reusable K/V
    memory, substantially reducing autoregressive inference latency.

    \item Extensive experiments on multiple LiDAR datasets demonstrate that
    SPEM achieves state-of-the-art performance for both lossy and lossless
    geometry compression, with particularly substantial bitrate reductions for
    lossless compression in the Cartesian coordinate space.
\end{itemize}

%% file: sec/2_Related_work.tex
\section{Related Work}
\label{sec:related_work}

\subsection{Voxel-based Point Cloud Compression}
Voxel-based methods quantize point cloud geometry into regular voxels and learn compact representations or occupancy distributions with 3D convolutional networks. Early approaches~\cite{quach2019learning,quach2020improved} used
dense 3D convolutions, but processing numerous empty voxels incurs substantial overhead. Sparse convolution~\cite{choy20194d,tang2022torchsparse,tang2023torchsparse++}
alleviates this issue by operating only on occupied locations.
Wang et al.~\cite{wang2021multiscale} introduced a multiscale sparse tensor
representation for progressive occupancy prediction, later extended by
SparsePCGC~\cite{wang2022sparse} with multiscale occupancy modeling.
Unicorn~\cite{wang2024versatile} further developed universal multiscale
conditional coding, while UniPCGC~\cite{wang2025unipcgc} introduced uneven
stage-wise occupancy coding and dynamic sparse convolution for flexible
rate--complexity control.
More recently,~\cite{liu2026voxel,you2025reno,huang2025linr,liu2026next}
reorganized spatial occupancy context into the channel dimension to enlarge the effective context and reduce autoregressive overhead.
ELiC~\cite{kim2026elic} improves efficiency through cross-bit-depth feature propagation and adaptive encoder selection while preserving Morton ordering. Despite these advances, voxel-based models mainly aggregate local voxel neighborhoods, with limited effective receptive fields at fine resolutions, particularly for high-precision LiDAR point clouds.

\subsection{Octree-based Point Cloud Compression}
Octree-based methods recursively partition 3D space and encode node occupancy, providing a compact hierarchical representation for LiDAR point clouds. The G-PCC standard~\cite{mpeg2023gpcc2} adopts octree coding as a fundamental geometry representation and employs hand-crafted context models to estimate occupancy probabilities from previously coded spatial information. Building upon this hierarchical representation, OctSqueeze~\cite{huang2020octsqueeze}
pioneered learned octree entropy modeling with ancestral context, followed by
VoxelContext-Net~\cite{que2021voxelcontext} using local voxel context and
OctAttention~\cite{fu2022octattention} exploiting sibling and ancestor dependencies through attention.
Chen et al.~\cite{chen2022point} further incorporated sibling context and local surface priors. EHEM~\cite{song2023efficient} introduced hierarchical attention and grouped context for efficient parallel decoding, while subsequent works~\cite{fan2023multiscale,jin2024ecm} further improved cross-node dependency modeling and parallel occupancy coding.
SCP~\cite{luo2024scp} transformed Cartesian geometry into spherical coordinates to explicitly exploit the LiDAR scanning prior. More recently, TopNet~\cite{wang2025topnet} combines local and long-range context, while PACE~\cite{zhu2026pace} decouples non-causal context aggregation from stage-scalable causal prediction.
In contrast, prior scan-prior-aware methods explicitly exploit the LiDAR scanning prior through coordinate transformation, which incurs precision loss and precludes strict lossless coding. SPEM instead can implicitly exploit the scanning prior directly in the Cartesian octree, preserving the original geometry without requiring coordinate
transformation.

%% file: sec/3_Method.tex
\section{Method}

\subsection{Overall Framework}

\begin{figure*}[thb]
    \centering
    \includegraphics[width=0.98\textwidth]{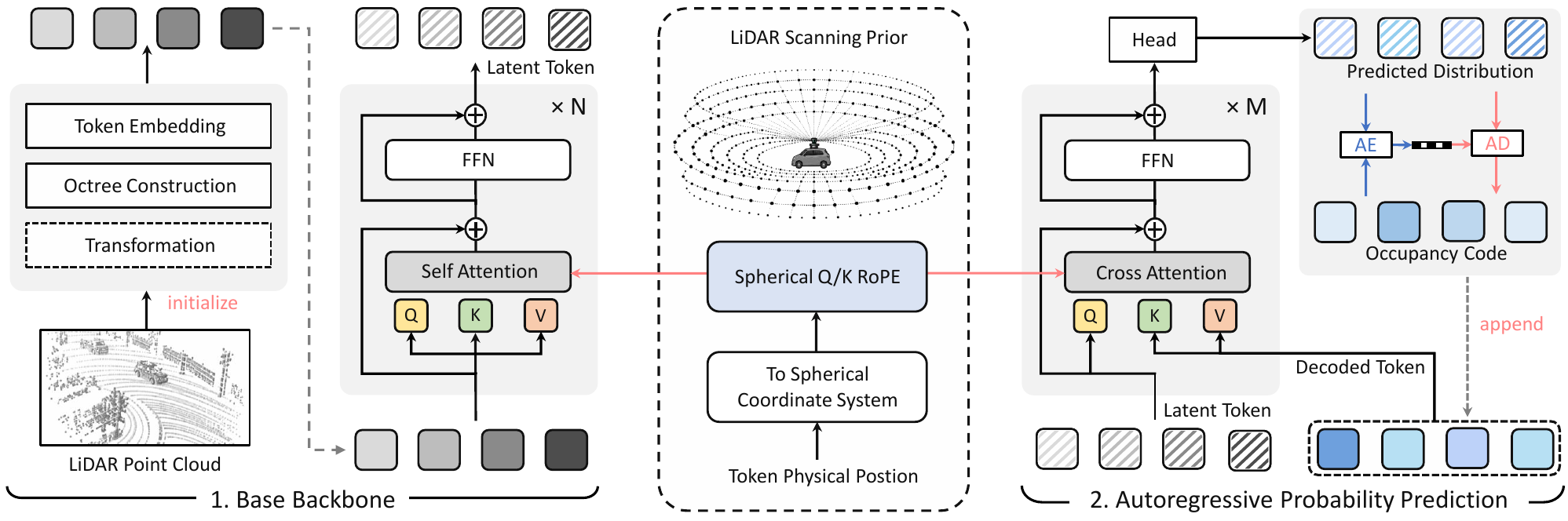}
    \caption{
        Overview of our LiDAR point cloud compression framework.
        The base backbone extracts latent tokens from the input LiDAR point cloud,
        while the autoregressive module predicts occupancy distributions conditioned
        on previously decoded tokens. We further incorporate LiDAR scanning priors
        through spherical Q/K RoPE based on physical coordinates.
    }
    \label{fig:framework}
\end{figure*}

As illustrated in Fig.~\ref{fig:framework}, SPEM first converts the input LiDAR point cloud into an octree-based token representation, models their spatial context with a scan-prior-aware backbone, and progressively estimates occupancy code through multi-stage autoregressive prediction for entropy coding.

(i) Token Initialization:
Given a LiDAR point cloud in Cartesian coordinates $(x,y,z)$, we optionally apply a scan-aligned transformation, such as spherical~\cite{luo2024scp} or cylindrical-beam coordinates~\cite{zhu2026pace}, before octree construction.
Note that SPEM is less sensitive to this transformation by modeling the scanning prior within the entropy model. We then partition the serialized octree nodes into local windows and aggregate
hierarchical context to form the input tokens. Details are provided in Sec.~\ref{sec:tokenization}.

(ii) Base Backbone:
Each local window of 1,024 tokens is fed into a Transformer encoder with $N=12$ layers, where each layer follows a pre-norm architecture with multi-head self-attention and a two-layer FFN.
To integrate the geometric priors of LiDAR scanning into the backbone, we introduce a scan-aware positional encoding into the self-attention operation. The resulting latent tokens capture spatial context aligned with the LiDAR scanning pattern for subsequent occupancy-code prediction. Details of the positional encoding are provided in Sec.~\ref{sec:positional_encoding}.

(iii) Autoregressive Probability Prediction:
For each latent token, its backbone feature serves as the query, while previously decoded nodes are encoded into memory keys and values using their backbone features together with occupancy and hierarchical information. The query is refined through $M=4$ cross-attention blocks, with our scan-aware positional encoding applied to both the query and memory keys. A prediction head finally outputs a 255-way occupancy distribution for entropy coding. Details of the multi-stage prediction scheme are provided in Sec.~\ref{sec:multi_stage}.

\subsection{Point Cloud Tokenization}
\label{sec:tokenization}

\textbf{Transformation.} We consider three coordinate representations before octree construction, following the setting in~\cite{zhu2026pace}, as summarized in Tab.~\ref{tab:coordinate_representation}.
Cartesian coordinates $(x,y,z)$ preserve the original lossless geometry, whereas spherical coordinates $(r,\theta,\phi)$ and cylindrical-beam coordinates $(\rho,\theta,b)$ better align with the LiDAR scanning pattern. However, their discretized scan-aligned representations do not exactly preserve the original Cartesian coordinates, resulting in coordinate precision loss and preventing strictly lossless reconstruction.

\begin{table}[t]
    \centering
    \caption{Coordinate representations before octree construction.}
    \label{tab:coordinate_representation}
    \small
    \setlength{\tabcolsep}{4.5pt}
    \renewcommand{\arraystretch}{1.15}
    \begin{tabular}{@{}cccc@{}}
        \toprule
        \textbf{Representation} &
        \textbf{Coordinates} &
        \textbf{Intrinsics} &
        \textbf{Lossless} \\
        \midrule
        Cartesian
        & $(x,y,z)$
        & $\times$
        & $\checkmark$ \\

        Spherical
        & $(r,\theta,\phi)$
        & $\times$
        & $\times$ \\

        Cylindrical-beam
        & $(\rho,\theta,b)$
        & $\checkmark$
        & $\times$ \\
        \bottomrule
    \end{tabular}
\end{table}

\noindent\textbf{Octree Construction}
The transformed point cloud is organized into a hierarchical octree with a maximum depth of $L$, where each non-empty node is represented by an 8-bit occupancy code $a_i^\ell \in \{1,\ldots,255\}$ indicating the occupancy states of its eight children. The octree nodes are traversed level by level, while nodes within each level
follow the Morton ordering induced by the parent order and child
octant indices. This yields an ordered occupancy sequence
$\mathcal{A}^\ell = \{a_1^\ell,\ldots,a_{N_\ell}^\ell\}$ at each level $\ell$ for tokenization.

\noindent\textbf{Token Embedding.}
The nodes at each octree level are partitioned into non-overlapping windows of at most 1,024 nodes. Each token combines three-level ancestor occupancy with the current octant and level embeddings, while the occupancy code is used only as the prediction target.
Following~\cite{zhu2026pace}, we employ graph-based positional encoding (GPE) to aggregate features from spatial neighbors.
However, GPE only captures local spatial relations without explicitly modeling the LiDAR scanning prior. 
 
\subsection{Spherical Positional Encoding}
\label{sec:positional_encoding}

\begin{figure}[t]
    \centering
    \includegraphics[width=\columnwidth]{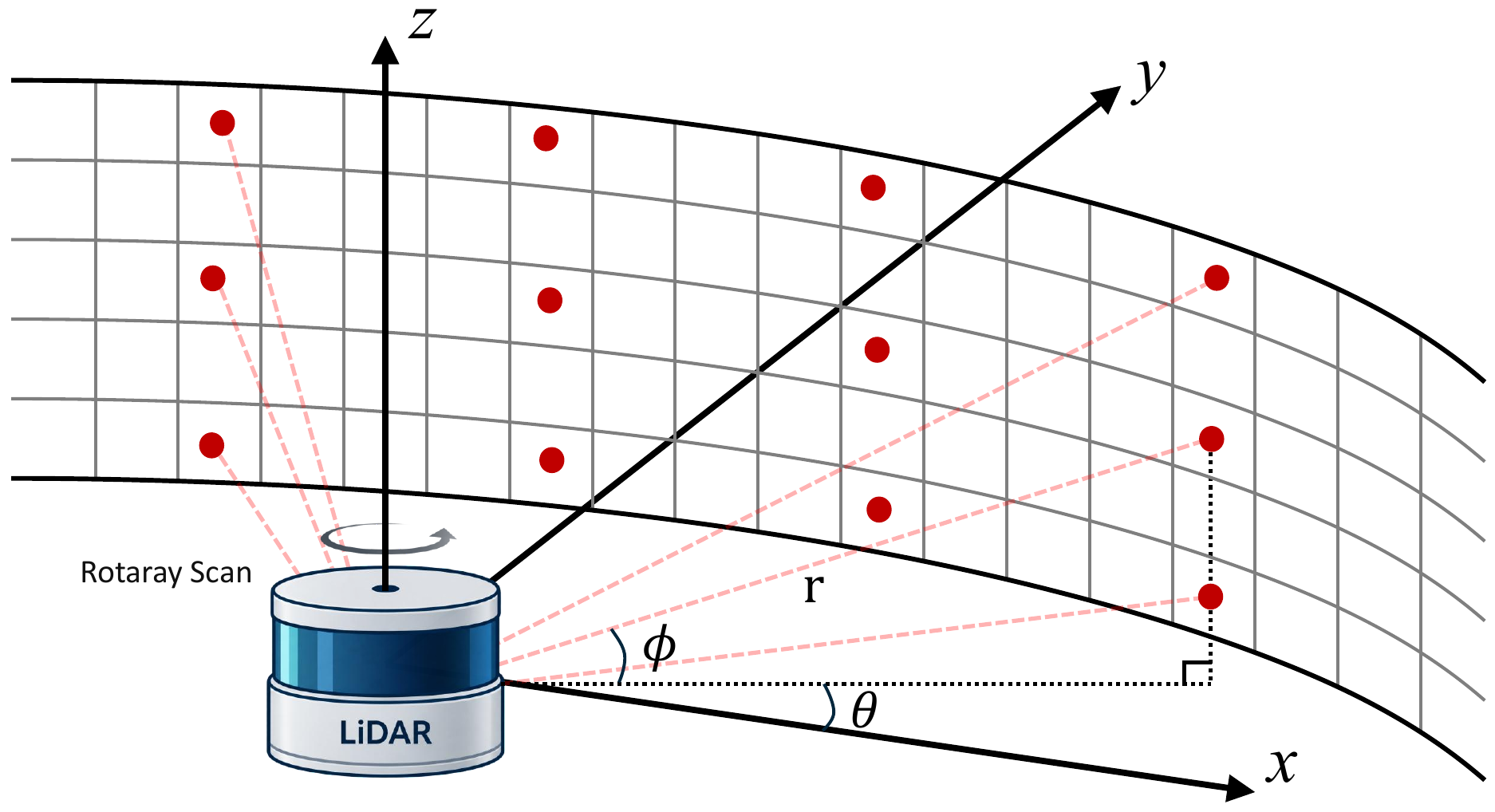}
    \caption{LiDAR rotating scanning prior for spherical positional encoding. Each token position is mapped from Cartesian space to spherical coordinates \((r,\theta,\phi)\), where \(r\), \(\theta\), and \(\phi\) denote range, azimuth, and elevation, respectively, providing a scan-aligned spatial description for positional encoding.}
    \label{fig:lidar_scanning_prior}
\end{figure}

As illustrated in Fig.~\ref{fig:lidar_scanning_prior}, the rotating scanning pattern of LiDAR is characterized in spherical coordinates. To incorporate this scanning geometry into entropy modeling, we map each token to its spherical position and construct multi-frequency rotary positional encoding for attention.

\noindent\textbf{Token Position Mapping.}
For each token in the Cartesian octree, let $D$ denote the stored tree depth
and $\ell$ its level, with $\ell=1$ for the root.
Let $\mathbf{p}_i$ denote its lower-corner grid coordinate,
$\mathbf{o}$ the physical lower corner of the root voxel in the LiDAR-centered
coordinate system, and $\boldsymbol{\delta}$ the finest-grid spacing.
Defining the center offset at level $\ell$ as
\begin{equation}
    \boldsymbol{\Delta}_{\ell}
    =
    \left(
    2^{D-\ell},
    2^{D-\ell},
    2^{D-\ell}
    \right),
\end{equation}
the physical center of token $i$ is
\begin{equation}
    \mathbf{c}_i =
    \mathbf{o} +
    \left(
    \mathbf{p}_i+\boldsymbol{\Delta}_{\ell}
    \right)
    \odot \boldsymbol{\delta}.
\end{equation}
Its spherical position is then obtained as
\begin{equation}
\begin{aligned}
    r_i &= \|\mathbf{c}_i\|_2, \\
    \theta_i &=
    \operatorname{atan2}(c_{i,y},c_{i,x}) \bmod 2\pi, \\
    \beta_i &=
    \frac{\pi}{2}-\phi_i
    =
    \arccos\left(\frac{c_{i,z}}{r_i}\right),
\end{aligned}
\end{equation}
where $r_i$ and $\theta_i$ denote the range and azimuth, while $\phi_i$
is the elevation illustrated in Fig.~\ref{fig:lidar_scanning_prior} and
$\beta_i$ is the corresponding polar angle.

The token position is fully determined by its grid coordinate, level, and
fixed root parameters before predicting the current occupancy code.
SPEM therefore retains the Cartesian octree for geometry coding and uses
spherical coordinates only for positional encoding, while the normalized
Cartesian positions used by GPE remain unchanged.
For the spherical and cylindrical-beam tree representations considered above,
the corresponding $(r,\theta,\beta)$ coordinates are obtained from the stored
node coordinates and quantization steps, or from calibrated mean pitch angles
over the decoded beam prefixes, respectively.

\noindent\textbf{Multi-Frequency Rotary Encoding.}
Given the spherical position $[r,\theta,\beta]$, we construct multi-frequency rotary phases for each 64-D attention head. The 32 two-dimensional rotation pairs are allocated to range, azimuth, and
polar angle using $7$, $16$, and $9$ pairs, respectively. Motivated by the anisotropic sampling pattern of rotating LiDAR, we allocate
more rotation pairs to azimuth, which is typically sampled more densely than the vertical direction and thus benefits from richer angular frequencies. The corresponding phases are defined as
\begin{equation}
    \gamma_k^{r}=\frac{2\pi r}{\lambda_k},
    \qquad
    \gamma_k^{\theta}=n_k\theta,
    \qquad
    \gamma_k^{\beta}=m_k\beta,
\end{equation}
where $\lambda_k$ denotes the physical wavelength for range, while $n_k$ and $m_k$ are angular harmonics for the azimuth and polar dimensions, respectively. The frequency allocation for each spherical dimension is summarized in
Tab.~\ref{tab:spherical_frequency}.

\begin{table}[t]
    \centering
    \caption{Frequency allocation for spherical positional encoding.}
    \label{tab:spherical_frequency}
    \fontsize{8.7pt}{10.2pt}\selectfont
    \setlength{\tabcolsep}{2.5pt}
    \renewcommand{\arraystretch}{1.10}

    \begin{tabularx}{\columnwidth}{
        @{}
        >{\centering\arraybackslash}m{0.24\columnwidth}
        >{\centering\arraybackslash}X
        @{}
    }
        \toprule
        \textbf{Component} &
        \textbf{Wavelengths / harmonics} \\
        \midrule

        Range $r$ (7)
        &
        \begingroup
        \setlength{\arraycolsep}{2pt}
        $\displaystyle
        \begin{array}{c@{\;}c}
            \lambda~(\mathrm{m}): &
            \left\{
            \begin{array}{ccccccc}
                0.5 & 1 & 2 & 4 & 8 & 16 & 32
            \end{array}
            \right\}
        \end{array}$
        \endgroup
        \\[2pt]

        Azimuth $\theta$ (16)
        &
        \begingroup
        \setlength{\arraycolsep}{2pt}
        $\displaystyle
        \begin{array}{c@{\;}c}
            n: &
            \left\{
            \begin{array}{cccccccc}
                1  & 2  & 4  & 8   & 12  & 16  & 24  & 32 \\
                48 & 64 & 96 & 128 & 192 & 256 & 512 & 1024
            \end{array}
            \right\}
        \end{array}$
        \endgroup
        \\[2pt]

        Polar $\beta$ (9)
        &
        \begingroup
        \setlength{\arraycolsep}{2pt}
        $\displaystyle
        \begin{array}{c@{\;}c}
            m: &
            \left\{
            \begin{array}{ccccccccc}
                1 & 4 & 8 & 16 & 32 & 64 & 128 & 256 & 512
            \end{array}
            \right\}
        \end{array}$
        \endgroup
        \\

        \bottomrule
    \end{tabularx}
\end{table}

For each two-dimensional channel pair, the corresponding query and key features
are rotated according to
\begin{equation}
    \mathcal{R}(\gamma)=
    \begin{bmatrix}
        \cos\gamma & -\sin\gamma \\
        \sin\gamma &  \cos\gamma
    \end{bmatrix}.
\end{equation}
This allows attention to capture relative spherical phase differences between
tokens and thereby model scan-aligned spatial relations.
In particular, the integer azimuth harmonics ensure periodic consistency across
the $\theta=0$ and $\theta=2\pi$ boundary.
The proposed spherical RoPE is applied to the Q/K features in both backbone
self-attention and decoded-memory cross-attention.

\subsection{Autoregressive Probability Modeling}
\label{sec:multi_stage}

\begin{figure}[t]
    \centering
    \includegraphics[width=\columnwidth]{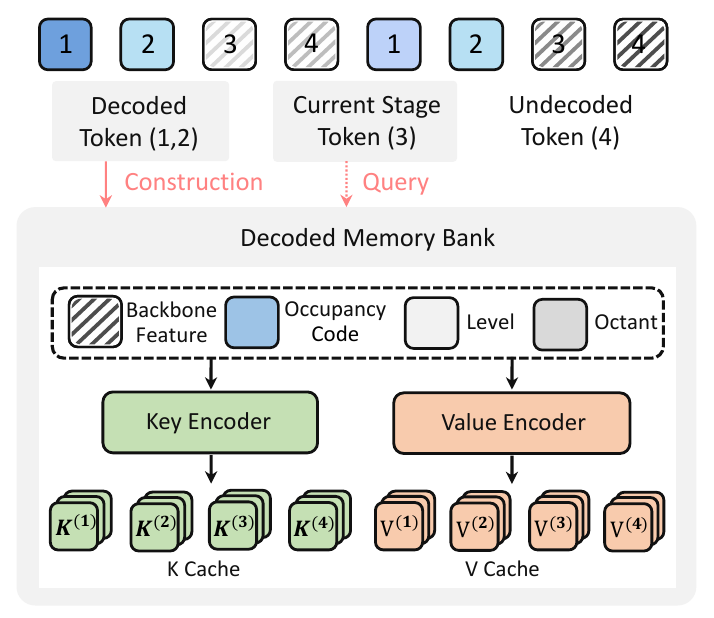}
    \caption{
    Stage-wise autoregressive probability modeling in SPEM.
    Previously decoded tokens are encoded into cached K/V memory, while current-stage
    tokens query the accumulated memory for occupancy prediction; newly decoded tokens
    are then appended to the memory for subsequent stages.
    }
    \label{fig:autoregressive_modeling}
\end{figure}

As illustrated in Fig.~\ref{fig:autoregressive_modeling}, SPEM adopts a
stage-wise autoregressive scheme to balance parallel probability prediction
and causal dependency modeling.
Within each window, tokens are interleaved into $S$ stages.
When predicting the current stage, occupancy codes from preceding stages have
already been decoded and encoded into a persistent memory bank, while tokens
in later stages remain inaccessible.

Let $\mathcal{I}_s$ denote the set of token indices assigned to stage $s$, and
$\mathcal{I}_{<s}=\bigcup_{u<s}\mathcal{I}_u$ denote those in preceding stages.
The occupancy distribution of a window is factorized as
\begin{equation}
    p(\mathbf{a}\mid\mathbf{F})
    =
    \prod_{s=1}^{S}
    \prod_{i\in\mathcal{I}_s}
    p\!\left(
        a_i
        \mid
        \mathbf{F},
        \mathbf{a}_{\mathcal{I}_{<s}}
    \right),
\end{equation}
where $\mathbf{F}$ denotes the backbone features and
$\mathbf{a}_{\mathcal{I}_{<s}}$ contains the occupancy codes decoded in preceding stages. For each token in the current stage, its backbone feature serves as the query to retrieve decoded context from the memory bank through cross-attention. Our spherical positional encoding is applied to the query and memory keys, enabling scan-aligned spatial modeling during memory retrieval. Tokens within the same stage are predicted in parallel, producing a 255-way probability distribution over occupancy codes for arithmetic encoding and decoding. After the stage is decoded, its occupancy codes are appended to the memory bank and become available to the subsequent stage.

\noindent\textbf{Decoded Memory Construction.}
For each previously decoded token, we combine its backbone feature with the decoded occupancy code and its level and octant information.
As shown in Fig.~\ref{fig:autoregressive_modeling}, two independent encoders transform this context into key and value representations.
Each prediction layer maintains its own projected key and value cache, denoted by $\{K^{(t)},V^{(t)}\}_{t=1}^{M}$. Once a token is decoded, its per-layer K/V representations are computed once
and cached for all subsequent stages. Since the memory construction is independent of the stage assignment, changing the number of stages only modifies the decoding schedule without introducing
stage-specific parameters, allowing our proposed SPEM to support flexible stage configurations.

%% file: sec/4_Experiment.tex
\section{Experiment}

\subsection{Experimental Settings}

\noindent\textbf{Dataset.}
We evaluate SPEM on four LiDAR point cloud datasets:
SemanticKITTI~\cite{behley2019iccv},
Ford~\cite{pandey2011ford},
nuScenes~\cite{caesar2020nuscenes},
and QNX~\cite{flynn2018qnx}.
For SemanticKITTI, sequences 00--10 are used for training and 11--21 for testing, yielding 23,201 training frames and 20,351 test frames. For Ford, sequence 01 is used for training and sequences 02--03 for testing, with each sequence containing 1,500 frames.
For nuScenes, we follow the data split in~\cite{zhu2026pace} and select 6,000 training frames and 450 test frames from non-overlapping scenes. For QNX, \emph{Motorway Bridge} and \emph{Navigating Turns} are used for training, while \emph{Junction Approach} and \emph{Junction Exit} are used for testing, yielding 1,111 training frames and 148 test frames. SemanticKITTI and nuScenes provide floating-point Cartesian coordinates in meters,
whereas Ford and QNX provide geometry on a native 1-mm integer grid.

\noindent\textbf{Baselines} We compare SPEM with traditional codecs and representative learning-based approaches. Traditional baselines include G-PCC TMC13v23~\cite{mpeg2023gpcc2}, octree and predictive modes, as well as L3C2~\cite{lasserre2021lidar}. Learning-based baselines include RENO~\cite{you2025reno}, SparsePCGC~\cite{wang2022sparse}, Unicorn~\cite{wang2024versatile}, OctAttention~\cite{fu2022octattention},  EHEM~\cite{song2023efficient}, SCP~\cite{luo2024scp}, TopNet~\cite{wang2025topnet}, ASRL~\cite{wang2025asrl}, LPCM~\cite{sun2026lpcm}, and PACE~\cite{zhu2026pace}. 
To evaluate different coordinate representations, we reproduce and retrain PACE with Cartesian, spherical, and cylindrical-beam inputs on selected datasets, achieving performance comparable to or better than the original reported results. For the other baselines, we use the results reported in their original papers.

\noindent\textbf{Metric}
We evaluate rate-distortion performance using bits per point (bpp) for bitrate and point-to-point (D1) and point-to-plane (D2) PSNR for geometry distortion. The PSNR peak values are set to 59.7~m for SemanticKITTI and nuScenes and
30,000~mm for Ford and QNX. We report R-D results under Cartesian, spherical, and cylindrical-beam representations to evaluate the influence of coordinate transformation.
We further use BD-BR~\cite{bjontegaard2001calculation} to measure the average bitrate difference at equivalent geometric quality, where lower values indicate better compression efficiency.
For Ford and QNX, we also evaluate lossless compression performance under Cartesian coding.

\noindent\textbf{Implementation settings}
We train a separate SPEM model for each dataset and coordinate representation. SPEM operates on 1,024-token windows with a 256-D feature dimension and four attention heads, using a 12-layer backbone and a 4-layer decoded-memory predictor with 1,024-D FFNs. During training, the autoregressive stage count is sampled from $S\in\{1,2,4,8\}$. We use $S=8$ for default R-D evaluation to favor compression performance,
while $S=1$ prioritizes inference speed. SPEM is optimized with cross-entropy loss on occupancy prediction using AdamW on NVIDIA A100 GPUs, with an initial learning rate of $5\times10^{-4}$ decayed by 0.8 approximately every 30,000 steps. SemanticKITTI, Ford, nuScenes, and QNX are trained for 5, 50, 25, and 20 epochs, respectively.

\begin{figure*}[thb]
      \centering
      \includegraphics[width=\textwidth]{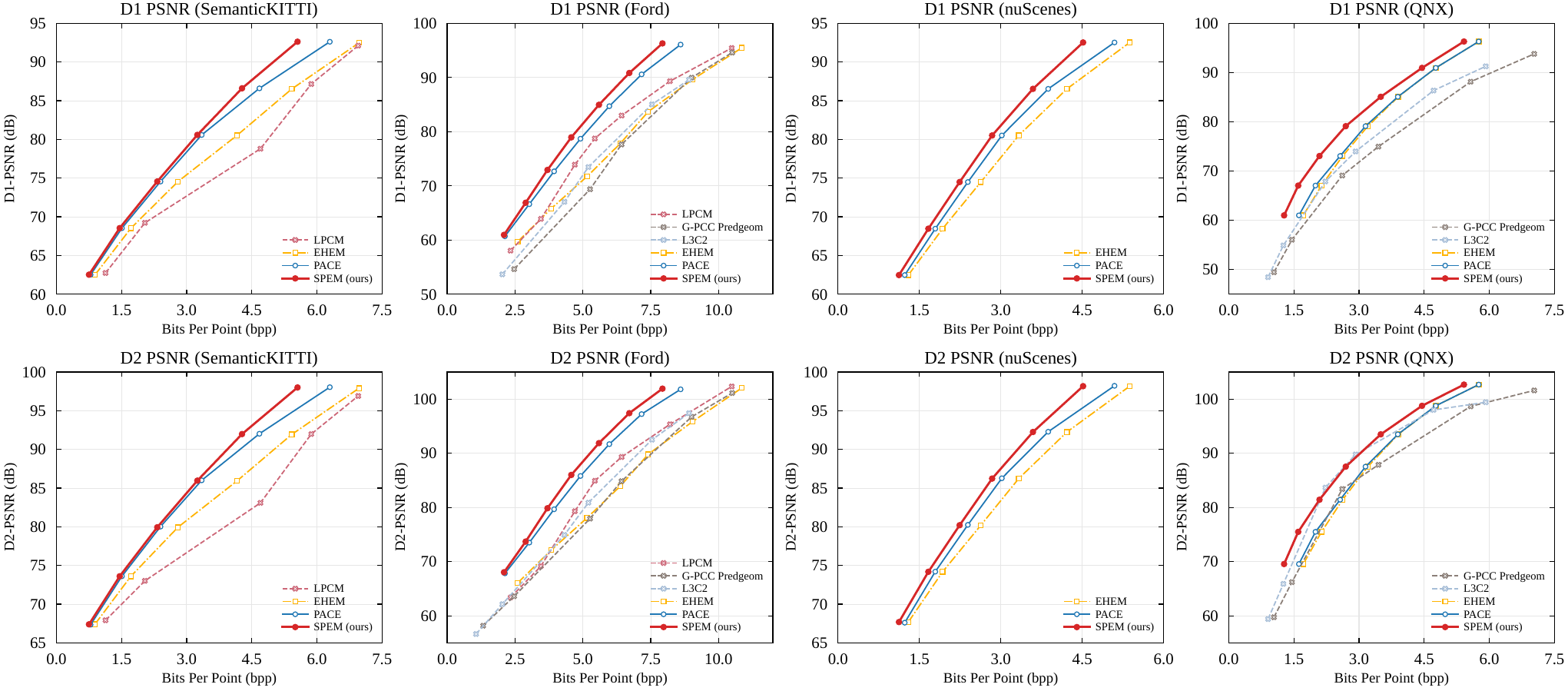}
      \caption{R-D comparison on SemanticKITTI, Ford, nuScenes, and QNX under coordinate transformation, using spherical coordinates for SemanticKITTI and nuScenes and cylindrical-beam coordinates for Ford and QNX.}
      \label{fig:rd_with_transform}
  \end{figure*}

  \begin{figure*}[thb]
      \centering
      \includegraphics[width=\textwidth]{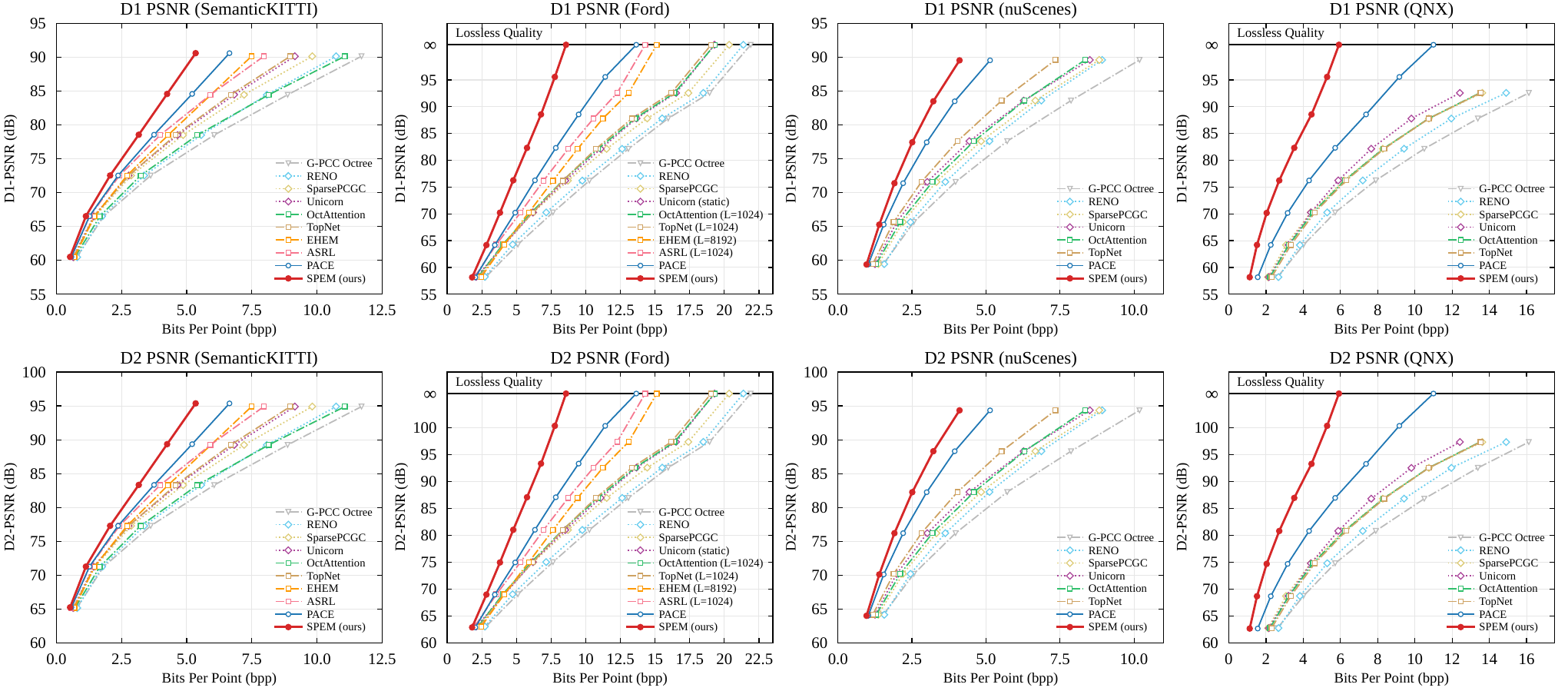}
      \caption{R-D comparison on SemanticKITTI, Ford, nuScenes, and QNX without coordinate transformation, using Cartesian coordinates; the $\infty$ markers for Ford and QNX indicate lossless compression at native millimeter precision.}
      \label{fig:rd_without_transform}
  \end{figure*}

\begin{table*}[t]
    \centering
    \caption{
    Comparison with PACE-AR under transformed and Cartesian representations. BD-rate is measured relative to PACE-AR, and encoding/decoding time is reported at the highest-rate operating point on each R-D curve.
    }
    \label{tab:spem_vs_pace}

    \footnotesize
    \setlength{\tabcolsep}{2.6pt}
    \renewcommand{\arraystretch}{1.18}

    \begin{tabular*}{\textwidth}{
        @{\extracolsep{\fill}}
        l
        cc cc
        cc cc
        ccc
        @{}
    }
        \toprule

        &
        \multicolumn{4}{c}{\textbf{Transformed}}
        &
        \multicolumn{4}{c}{\textbf{Cartesian}}
        &
        \multicolumn{3}{c}{\textbf{Lossless}} \\
        \cmidrule(lr){2-5}
        \cmidrule(lr){6-9}
        \cmidrule(l){10-12}

        \textbf{Dataset}
        & \makecell{BD-rate\\D1}
        & \makecell{BD-rate\\D2}
        & \makecell{Time\\PACE-AR}
        & \makecell{Time\\SPEM-S8}

        & \makecell{BD-rate\\D1}
        & \makecell{BD-rate\\D2}
        & \makecell{Time\\PACE-AR}
        & \makecell{Time\\SPEM-S8}

        & \makecell{PACE-AR\\(bpp)}
        & \makecell{SPEM-S8\\(bpp)}
        & Gain \\
        \midrule

        SemanticKITTI
        & \textbf{-5.30\%}
        & \textbf{-5.06\%}
        & 12.82/18.15
        & 2.19/2.25

        & \textbf{-14.53\%}
        & \textbf{-14.52\%}
        & 10.74/15.41
        & 2.02/2.07

        & --
        & --
        & -- \\

        Ford
        & \textbf{-6.77\%}
        & \textbf{-6.53\%}
        & 17.05/23.91
        & 2.60/2.59

        & \textbf{-24.01\%}
        & \textbf{-24.00\%}
        & 16.00/22.45
        & 2.63/2.56

        & 13.643
        & \textbf{8.576}
        & \textbf{37.14\%} \\

        nuScenes
        & \textbf{-7.21\%}
        & \textbf{-7.12\%}
        & 8.84/12.49
        & 1.09/1.22

        & \textbf{-14.28\%}
        & \textbf{-14.23\%}
        & 7.71/10.84
        & 1.05/1.14

        & --
        & --
        & -- \\

        QNX
        & \textbf{-14.19\%}
        & \textbf{-14.62\%}
        & 11.32/15.78
        & 1.16/1.28

        & \textbf{-36.76\%}
        & \textbf{-36.78\%}
        & 10.25/14.32
        & 1.21/1.30

        & 10.998
        & \textbf{5.920}
        & \textbf{46.17\%} \\

        \bottomrule
    \end{tabular*}
\end{table*}

\subsection{Performance Evaluation}
Fig.~\ref{fig:rd_with_transform} compares the R-D performance under scan-aligned
coordinate transformations, where spherical coordinates are used for
SemanticKITTI and nuScenes, while cylindrical-beam coordinates are adopted for
Ford and QNX. SPEM consistently achieves better R-D performance than the
compared methods for both D1 and D2. As quantified in Tab.~\ref{tab:spem_vs_pace}, SPEM achieves BD-rate savings
of 5.30\%--14.19\% for D1 and 5.06\%--14.62\% for D2 relative to PACE-AR,
with average savings of 8.37\% and 8.33\%, respectively. These gains show that scan-aware positional modeling in attention and efficient stage-wise autoregressive prediction remain complementary to explicit scan-aligned coordinate transformation.

The advantage becomes substantially larger when operating directly on the
original Cartesian representation, as shown in Fig.~\ref{fig:rd_without_transform}. SPEM achieves D1 BD-rate savings of 14.53\%, 24.01\%, 14.28\%, and 36.76\% on SemanticKITTI, Ford, nuScenes, and QNX, respectively, with closely matched improvements in D2. Averaged over the four datasets, the gains increase to 22.40\% in D1 and 22.38\% in D2, compared with about 8.4\% under coordinate transformation. The consistently larger improvement in Cartesian space indicates that SPEM can capture the LiDAR scanning structure directly during entropy modeling, reducing the dependence on explicit scan-aligned coordinate transformations while preserving the original geometry representation.

The Cartesian setting further enables lossless evaluation on Ford and QNX at their native millimeter precision. As reported in
Tab.~\ref{tab:spem_vs_pace}, SPEM reduces the lossless bitrate from
13.643 to 8.576 bpp on Ford and from 10.998 to 5.920 bpp on QNX,
corresponding to savings of 37.14\% and 46.17\%, respectively, over PACE-AR. This highlights a key advantage of SPEM: it exploits LiDAR scanning regularities within the entropy model while preserving the original Cartesian geometry, achieving substantially more compact lossless coding without coordinate transformation.

\begin{table}[t]
\centering
\caption{Rate--complexity trade-off on Ford under lossless coding.}
\label{tab:ford_lossless_stages}

\footnotesize
\setlength{\tabcolsep}{3.2pt}
\renewcommand{\arraystretch}{1.05}

\begin{tabular*}{\columnwidth}{
@{\extracolsep{\fill}}
c
c
c
c
c
@{}
}
\toprule
\textbf{Method}
& \textbf{Stage}
& \textbf{Enc. (s)}
& \textbf{Dec. (s)}
& \textbf{Rate (bpp)} \\
\midrule

\multirow[c]{6}{*}{\shortstack[c]{PACE\\(10.64M)}}
& $S=1$  & 1.49  & 1.63  & 14.576 \\
& $S=2$  & 1.56  & 1.50  & 14.219 \\
& $S=4$  & 1.66  & 1.53  & 14.013 \\
& $S=8$  & 1.84  & 1.74  & 13.890 \\
& $S=16$ & 2.46  & 2.48  & 13.815 \\
& AR     & 16.00 & 22.45 & 13.643 \\
\midrule

\multirow[c]{4}{*}{\shortstack[c]{\textbf{SPEM}\\(13.88M)}}
& $S=1$ & 1.82 & 1.92 & 9.214 \\
& $S=2$ & 1.98 & 1.91 & 8.820 \\
& $S=4$ & 2.18 & 2.09 & 8.657 \\
& $S=8$ & 2.63 & 2.56 & \textbf{8.576} \\
\bottomrule
\end{tabular*}
\end{table}

\subsection{Complexity Analysis}

Tab.~\ref{tab:ford_lossless_stages} compares the rate--complexity trade-off between PACE and SPEM under different autoregressive stage counts.
At $S=1$, both methods perform fully parallel prediction.
SPEM is slightly slower than PACE because no real decoded memory is available, yet the decoded-memory predictor still executes all four cross-attention blocks, where each layer attends to the learned null K/V and is followed by an FFN with residual updates.\par
As the stage count increases, however, SPEM runtime grows only moderately while the bitrate consistently decreases. This favorable scaling is enabled by our memory-based autoregressive design: the K/V representations of decoded tokens are constructed once and cached for reuse in all subsequent stages. Therefore, increasing $S$ mainly changes the decoding schedule and progressively exposes more decoded context, without repeatedly recomputing historical
representations. For example, increasing $S$ from 1 to 8 substantially improves the lossless rate while introducing only a modest increase in encoding and decoding time. This allows SPEM to flexibly trade inference speed for compression efficiency using a single model.

\begin{figure}[t]
    \centering
    \includegraphics[width=\columnwidth]{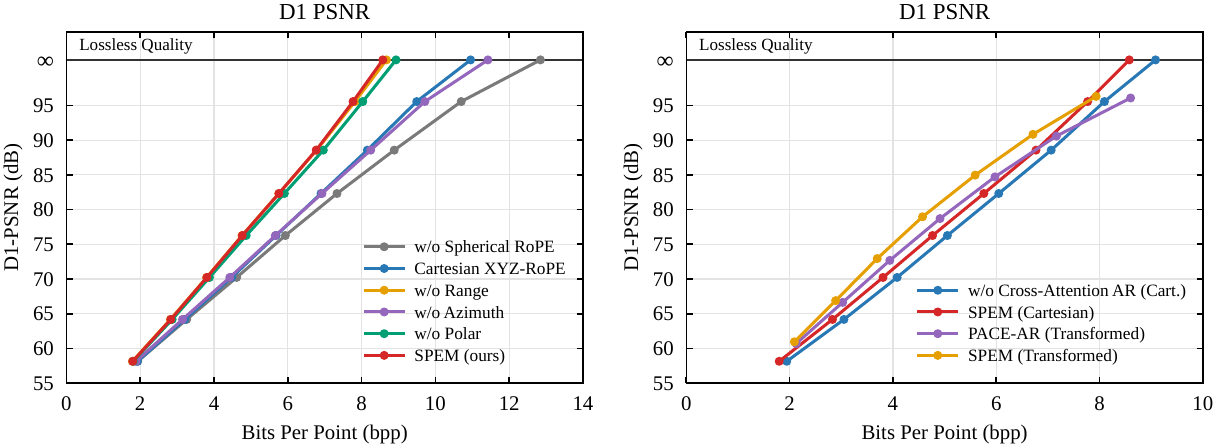}
    \caption{
    R-D comparison of different SPEM variants on Ford dataset.
    }
    \label{fig:ablation}
\end{figure}

\subsection{Qualitative Results}

\begin{figure*}[thb]
  \centering
  \includegraphics[width=\textwidth]{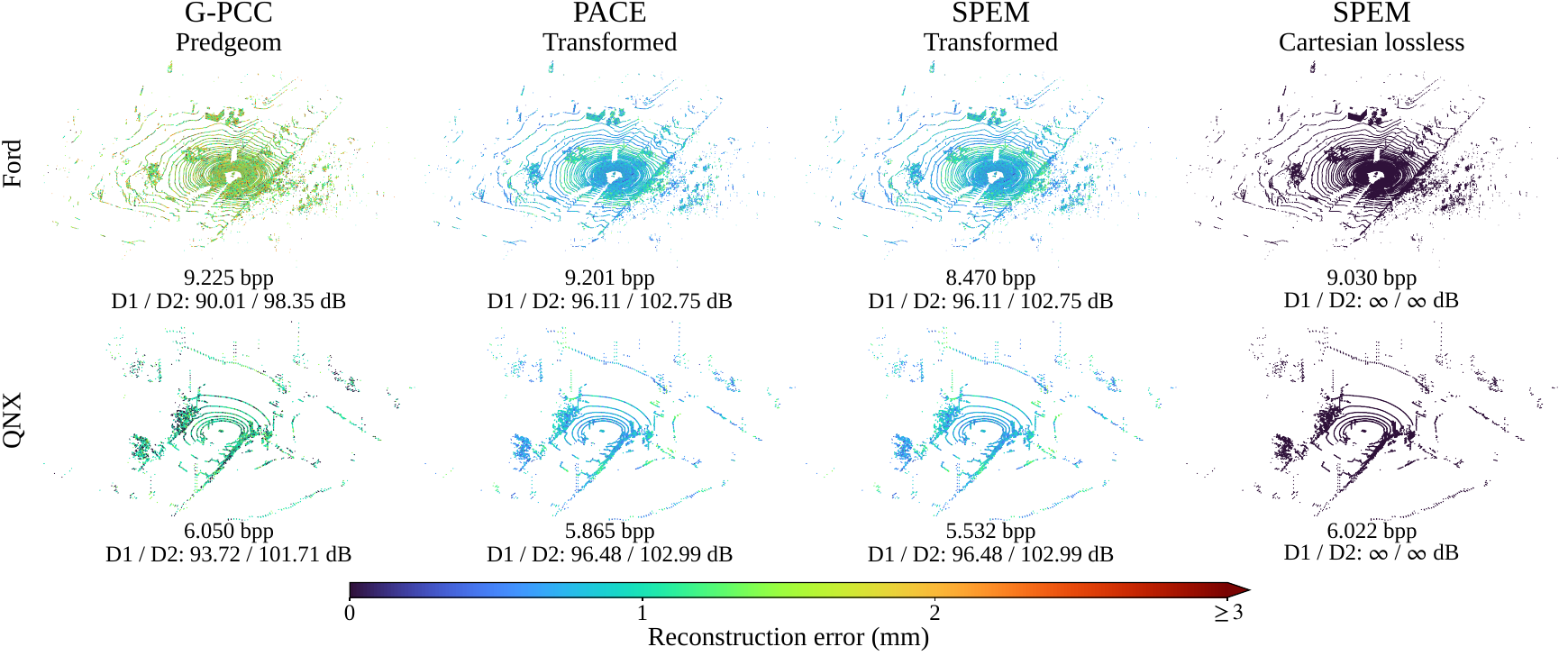}
  \caption{Qualitative comparison on Ford and QNX.
    From left to right, we show G-PCC predictive geometry, transformed PACE,
    transformed SPEM, and lossless Cartesian SPEM.
    Points are colored by reconstruction error in millimeters.}
  \label{fig:qualitative}
\end{figure*}

Fig.~\ref{fig:qualitative} visualizes reconstructions on Ford and QNX.
Compared with G-PCC predictive geometry, the learned methods produce
noticeably smaller geometric errors, while SPEM further improves coding
efficiency over PACE under the transformed representation.
Interestingly, for the transformed methods, reconstruction errors in distant
regions can be comparable to or even smaller than those in nearby regions.
This behavior results from the region-adaptive coding strategy inherited from~\cite{luo2024scp}, which assigns more octree levels to distant regions to compensate for their lower spatial sampling density. More importantly, Cartesian SPEM exactly preserves the original
millimeter-precision geometry at the lossless operating point.
These results show that SPEM can exploit LiDAR scanning regularities
while supporting both lossy and lossless geometry compression.

\subsection{Ablation Study}

\begin{figure}[t]
    \centering
    \includegraphics[width=\columnwidth]{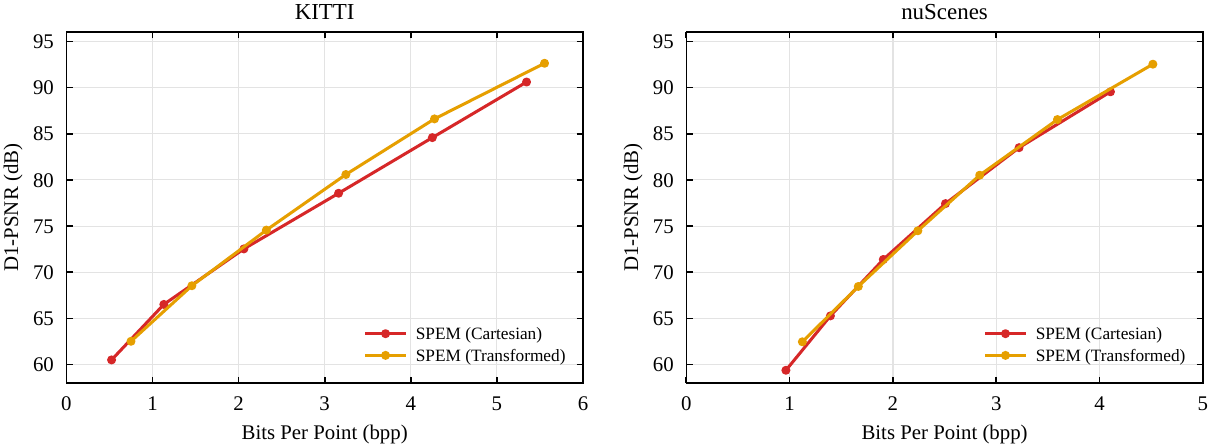}
    \caption{
    R-D comparison of SPEM with and without coordinate transformation on SemanticKITTI and nuScenes. 
    }
    \label{fig:ablation2}
\end{figure}

\noindent\textbf{Spherical positional encoding.}
The left column of Fig.~\ref{fig:ablation} evaluates the contribution of our
spherical positional encoding. Removing spherical RoPE causes a clear
degradation. Replacing it with Cartesian XYZ-RoPE partially
recovers the performance but remains notably inferior to SPEM, indicating that
the gain does not simply arise from rotary positional encoding itself, but from
aligning positional relations with the LiDAR scanning geometry.
Among the three spherical components, the relatively small degradation from
removing range suggests that the two angular components already capture the
dominant LiDAR scanning pattern. In addition, removing the azimuth phase causes
a larger performance drop than removing the polar phase, which is consistent
with the denser horizontal angular sampling of rotating LiDAR and motivates
allocating more frequency capacity to the azimuth dimension.

\noindent\textbf{Autoregressive predictor.}
The right column of Fig.~\ref{fig:ablation} evaluates the decoded-memory prediction design. Replacing our cross-attention predictor with the PACE-style causal predictor consistently degrades the R-D performance, including the lossless compression. This confirms the benefit of querying cached decoded context through cross-attention, which is further enhanced by our spherical RoPE to capture scan-aware spatial relations during memory retrieval.

\noindent\textbf{Effect of coordinate transformation.}
We investigate the necessity of explicit coordinate transformation by comparing SPEM under Cartesian and spherical representations, as shown in Fig.~\ref{fig:ablation2}. For a fair comparison, both configurations employ the same spherical positional encoding without using LiDAR intrinsic parameters. 
Their comparable R-D performance suggests that scan-aware entropy modeling effectively captures LiDAR scanning regularities without transforming the coded geometry. This indicates that the benefits of scan-aligned coordinates can largely be realized through probability modeling rather than changes to geometry representation. By decoupling acquisition priors from geometry representation, SPEM reduces reliance on explicit transformation while retaining Cartesian coding. This offers a promising direction toward unified lossy and lossless geometry compression.

%% file: sec/5_Conclusion.tex
\section{Conclusion}
\label{sec:conclusion}

We present SPEM, a scan-prior entropy modeling framework for
LiDAR point cloud geometry compression. Instead of relying solely on explicit
coordinate transformation to expose the LiDAR acquisition structure, SPEM
incorporates the scanning prior directly into entropy modeling through positional encoding and stage-wise autoregressive prediction.
The proposed spherical RoPE enables attention to model scan-aligned spatial
relations, while the decoded-memory predictor efficiently exploits previously decoded occupancy information through reusable K/V caches.
Experiments on multiple datasets demonstrate that SPEM achieves state-of-the-art performance for both lossy and lossless compression with favorable coding efficiency.